\def\year{2020}\relax
\documentclass[letterpaper]{article} 
\usepackage{aaai20}  
\usepackage{times}  
\usepackage{helvet} 
\usepackage{courier}  
\usepackage[hyphens]{url}  
\usepackage{graphicx} 

\usepackage{amsmath}
\usepackage{booktabs}
\usepackage{amsfonts}
\usepackage{amssymb}
\usepackage{bm}
\usepackage{xfrac}
\usepackage{framed}
\usepackage{enumitem}
\usepackage{mdframed}
\usepackage{ntheorem}

\usepackage{algorithm}
\usepackage{algorithmic}

\newcommand{\citet}[1]{\citeauthor{#1} \shortcite{#1}}

\usepackage{graphicx}  
\title{The Stanford Acuity Test: A Precise Vision Test Using \\
Bayesian Techniques and a Discovery in Human Visual Response}

\author{
Chris Piech,\thanks{Equal contribution.}$^1$
Ali Malik,$^{*1}$
Laura M Scott,
Robert T Chang,$^2$
Charles Lin$^2$  \\
$^1$Department of Computer Science, Stanford University \\
$^2$ Department of Ophthalmology, Stanford University \\
\{piech, malikali\}@cs.stanford.edu,
\{rchang3, lincc\}@stanford.edu
}

\begin{document}

\maketitle

\begin{abstract}
Chart-based visual acuity measurements are used by billions of people to diagnose and guide treatment of vision impairment.  However, the ubiquitous eye exam has no mechanism for reasoning about uncertainty and as such, suffers from a well-documented reproducibility problem. In this paper we make two core contributions. First, we uncover a new parametric probabilistic model of visual acuity response based on detailed measurements of patients with eye disease. Then, we present an adaptive, digital eye exam using modern artificial intelligence techniques which substantially reduces acuity exam error over existing approaches, while also introducing the novel ability to model its own uncertainty and incorporate prior beliefs. Using standard evaluation metrics, we estimate a 74\% reduction in prediction error compared to the ubiquitous chart-based eye exam and up to 67\% reduction compared to the previous best digital exam. For patients with eye disease, the novel ability to finely measure acuity from home could be a crucial part in early diagnosis. We provide a web implementation of our algorithm for anyone in the world to use. The insights in this paper also provide interesting implications for the field of psychometric Item Response Theory.
\end{abstract}

\section{Introduction}

Reliably measuring a person's visual ability is an essential component in the detection and treatment of eye diseases around the world. However, quantifying how well an individual can distinguish visual information is a surprisingly difficult task---without invasive techniques, physicians rely on chart-based eye exams where patients are asked visual questions and their responses observed.

Historically, vision has been evaluated by measuring a patient's \textit{visual acuity}: a measure of the font size at which a patient can correctly identify  letters shown a fixed distance away.  Snellen, the traditional eye exam chart, determines this statistic by marching down a set of discrete letter sizes, asking the patient a small number of questions per size to indentify the size where the patient gets less than half the letters correct. This approach is simple and is used daily in the treatment of patients; yet, it suffers from some notable shortcomings. Acuity exams such as these exhibit high variance in their results due to the large role that chance plays in the final diagnosis, and  the approximation error incurred by the need to discretise letter sizes on a chart. On the other hand, digital exams can show letter of any size and can \emph{adaptively} make decisions based on intelligent probabilistic models. As such they have  potential to address the shortcomings of analog charts.

While promising, contemporary digital exams have yet to dramatically improve accuracy over traditional chart-based approaches. The current best digital exam uses a psychometric Item Response Theory (IRT) algorithm for both selecting the next letter size to query and for making a final prediction of acuity. Under simulation analysis, this digital exam results in a 19\% reduction in error over traditional chart-based approaches. The separate fields of reinforcement learning and psychometric IRT have independently explored how to effectively make decisions under uncertainty. By merging the good ideas from both disciplines we can develop a much better visual acuity test.

In this paper we make two main contributions. First, we revisit the human Visual Response Function---a function relating the size of a letter to the probability of a person identifying it correctly---and discover that it follows an interpretable parametric form that fits real patient data. Second, we present an algorithm to measure a person's acuity which uses  several Bayesian techniques common in modern artificial intelligence. The algorithm, called the Stanford Acuity Test (StAT)\footnote{The previous state-of-the-art, FrACT, was named after Freiburg, the city in which it was developed. We continue in this tradition.}, has the following novel features:

\begin{enumerate}
\item Uses the new parametric form of the human Visual Response Function.
\item Returns a soft inference prediction of the patient's acuity, enabling us to express a calibrated confidence in the final result.
\item Uses a posterior probability matching algorithm to adaptively select the next letter size shown to a user. This effectively balances exploration of the acuity belief.
\item Accepts a patient's prior belief of their acuity, or alternatively, traces their vision over time.
\item Incorporates ``slip" estimation for unintended mistakes in the eye test process.
\end{enumerate}
We demonstrate that each of these additions lead to a more precise acuity exam.  In unison, the result is a test that is 74\% more accurate than the analog chart. Compared to the previous best digital exam, our experiments show an error reduction of up to 67\%.

For patients with more serious eye disease, the novel ability to finely measure acuity from home could play a crucial role in early diagnosis and effective treatment. We provide a web implementation for anyone in the world to use at \url{https://myeyes.ai}

\subsection{Visual Acuity}
Visual acuity is a measurement that captures a patient's visual ability in a succinct manner. It is defined to be the letter size at which the patient can correctly identify the displayed optotype (letter) with probability $\tau$, where the constant $\tau$ is set for each exam type.

\subsection{Chart-based Acuity Exams}

\begin{figure}
    \centering
    \includegraphics[width=\linewidth]{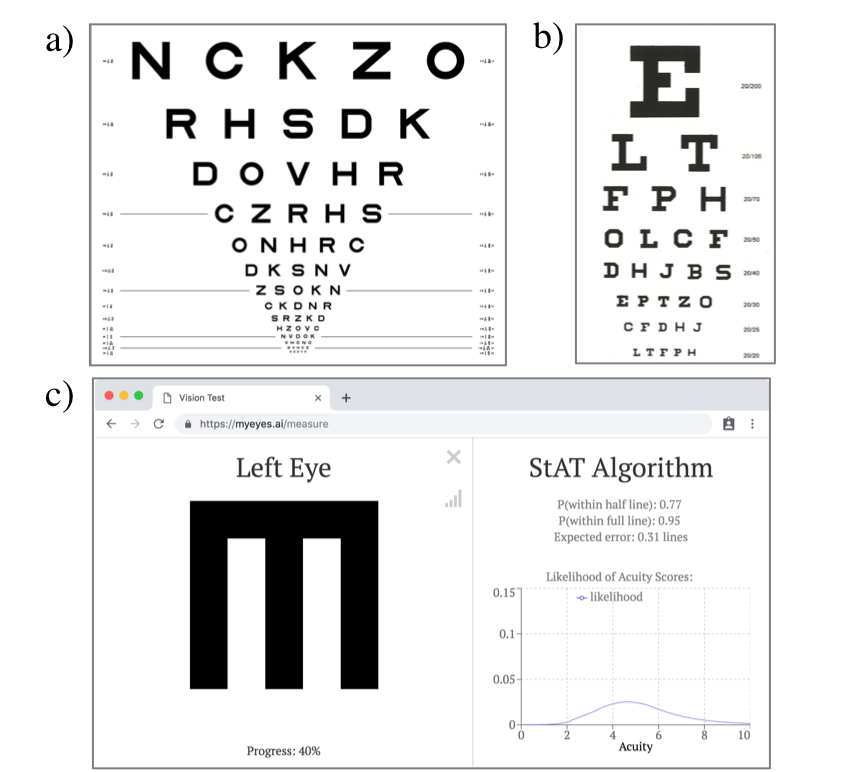}
    \caption{a) ETDRS, b) Snellen and c) StAT eye exams.}
    \label{fig:acuity_exams}
\end{figure}

In 1862, Herman Snellen developed the ubiquitous eye exam: a chart is placed at 6 meters (20ft) from the patient as they attempt to identify optotypes (letters) of progressively smaller sizes written on different lines (see Fig. \ref{fig:acuity_exams}). The goal is to find the optotype size at which the user can no longer identify at least half of the letters on the line. To keep the exam a reasonable duration, there is a small, discrete set of lines that are substantially different in size. The Snellen chart continues to be the most common acuity test, but there are other charts such as LogMar ETDRS Chart \cite{national1980recommended,ferris1982new},  Tumbling-E, Lea, and HOTV that generally use the same procedure with a different set of optotypes \cite{rosser2001development}. They all share the same core limitations:


\vspace{0.5em} \noindent \textbf{Guessing.\ } Making guesses is a critical part of an acuity exam. As the patient progresses to smaller optotypes, there is a gradual decrease in how easy it is to identify the letters. As optotype size decreases, the probability of correctly guessing decreases. This has the predictable problem that chance plays a large role in the final acuity score. As a concrete example, imagine an optotype size where the patient has a 0.5 probability of correctly identifying a letter. After guessing for five letters, the Binomial Theorem tells us there is still a 50\% chance that they incorrectly ``pass" the current line by getting more than 2 out of 5 correct.

\vspace{0.5em} \noindent \textbf{Discretisation. \ } Because of the limits of printing, it is necessary to discretise acuity scores in printed eye-chart exams. This makes it hard to have an acuity measure more precise than the predetermined discretisation. Discretisation is particularly limiting for patients who need to detect a small decreases in vision, as such a change could be indicative of a time sensitive need for medical intervention.

\vspace{0.5em} \noindent \textbf{Confidence. } Another limitation of all current tests is their inability to articulate their confidence  in the final measured acuity. Contemporary eye exams result in a ``hard" number for their acuity prediction as opposed to a ``soft" probability distribution. As an example, a soft prediction can make claims such as, ``there is a 75\% chance that the patient's true vision is within one line of our predicted acuity score." Current tests can only say how many letters were missed on the last line, but don't provide  probabilistic uncertainty.

Despite these limitations, chart-based acuity exams are used every day around the world to diagnose disease, inform treatment decisions, and evaluate medical research into new medications and best practices.

\subsection{Digital Acuity Challenge}

Computers enable digital, adaptive eye exams. A digital acuity eye exam proceeds as follows: the computer presents an optotype of a chosen font size, and the user guesses it (either correct or incorrect). The computer then gets to incorporate that response and chose a new font size to present. The test continues until either a fixed number of letters have been shown or till the model has determined an acuity score with sufficient confidence. A digital exam has two potential advantages over chart-based exams: (1) a computer can draw optotypes of any continuous size, and (2) a computer can adaptively choose the next letter size to show. The digital acuity challenge is to develop a policy for a digital eye exam that can hone in on a patient's true acuity statistic as fast and as accurately as possible.


\subsection{Prior Work}

The current state-of-the-art digital optotype size discrimination exam, the Freiburg Acuity Contrast Test (FrACT), was first developed in 1996 and has been successfully used in medical contexts since its conception \cite{bach1996freiburg,bach2006freiburg}.

FrACT builds an underlying model of human visual acuity which assumes that the probability a human correctly identifies a letter is a function, $v(x, v_0, s)$, of the letter size $x$ (see Appendix \ref{apdx:units} for discussion of units) and two parameters that change from person to person ($v_0$ and $s$):
\begin{equation}
    v(x, v_0, s) = c + (1-c)  / (1 + (v_0 \cdot x)^s).
\end{equation}

Here, $c$ is the probability that the human randomly guesses a letter correctly. The algorithm maintains a Maximum Likelihood Estimate (MLE). When choosing a next item size, the FrACT selects the size at which the function has the highest slope. The authors graciously offered to provide the algorithm source code. Digital exams, like FrACT, work especially well for patients with low vision \cite{schulze2006visual}.

More broadly, the FrACT test can be shown to reduce to Birnhaum's popular 3PL model which is the basis for Item Response Theory (IRT) literature used in psychometrics \cite{birnbaum1968some}. From an IRT perspective, each letter shown to a patient is an ``item" and the goal of the test is to uncover the latent ``ability" of a patient to see optotypes, whose ``size" is a function of their difficulty. The 3PL model makes the same logistic assumption for the relationship between difficulty of an item and probability of answering correctly that is made by the FrACT algorithm. As such, the improvements that we develop for the digital eye exams outlined in this paper are likely to be relevant for the many applications of IRT in testing beyond Ophthalmology.

\section{Human Visual Acuity Curve}

A central assumption of a vision exam is the function which relates the size of a letter, $x$, to the probability that the patient correctly identifies it: $v(x)$. This function is called the Visual Response Function (VRF)  \cite{bach1996freiburg}.

\begin{figure}[!t]
\centering
\includegraphics[width=0.95\linewidth]{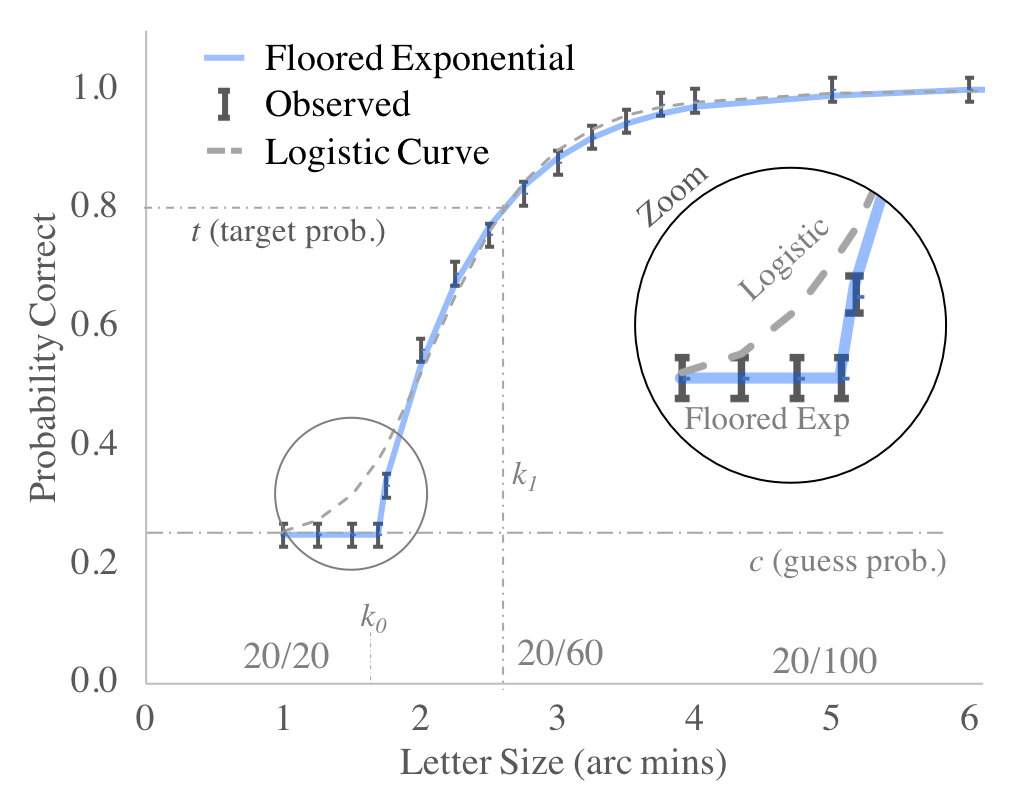}
\caption{An example of a single person's visual response function. The FrACT model uses a logistic which is inaccurate for low probabilities. Error bars are Beta distribution standard deviation after $> 500$ measurements.}
\label{flooredFit}
\end{figure}

Given enough patience, one can precisely observe the VRF for a single human.  This is very different from a typical psychometric exam where it would be unreasonable to ask a patient the same question hundreds of times. Previous studies have measured VRF curves and concluded that they are best fit by a logistic function, an assumption that was adopted by FrACT \cite{petersen1990fehlerbreite}. However, this previous work missed the notable deviation of the logistic function from human response close to the point where a patient transitions from random guessing to actually discerning visual information (see Fig. \ref{flooredFit}).


We conducted an IRB-approved experiment at the Stanford Byers Eye Institute and carefully measured the Visual Response Function for patients with different vision-limiting eye diseases. Twelve patients were shown randomly selected optotypes of a fixed size until we were confident about their probability of responding correctly at that size. We represented our uncertainty about this probability at each optotype size as a Beta distribution (the conjugate prior of the Bernoulli) and continued testing until our uncertainty about the probability was below a fixed threshold. Patients took breaks between questions and we randomized the order of letter size to remove confounds such as tear film dryness that can lead to vision decrease over the course of an exam. Surprisingly, we found that the traditional assumption for the VRF---a logistic curve---struggled to fit the responses we collected, especially at small letter sizes where the subjects were randomly guessing.

Figure \ref{flooredFit} shows an example of a single patient who volunteered to answer over 500 answers to optotypes questions of varying sizes (see Appendix \ref{apdx:fexp_vrfs} for the remaining patients). These experiments depict the VRF as a combination of two processes: for tiny letters a patient is unable to see and guesses randomly, with $v(x) = c$. For letters where the patient can discern information, their probability of answering correctly is determined by an exponential function, traditionally parameterised by location $b$ and scale $\lambda$ i.e. $v(x) = 1 - e^{-\lambda(x - b)}$. The mixture of these two processes is a ``Floored Exponential'' function. The resulting equation can be reparameterised with values that eye care providers find meaningful (see Appendix \ref{apdx:fexp_reparam} for derivation):

\begin{framed}

\noindent \textbf{Floored Exponential}

\noindent A Floored Exponential is a maximum between a constant floor and an exponential function. For visual acuity we parameterise it as:
\begin{equation*}
     v(x, k_0, k_1) = \text{max}\left\{c, \  1 - (1 - c)\left(\frac{1 - \tau}{1-c}\right)^{\frac{x - k_0}{k_1 - k_0}}\right\},
\end{equation*}
where $x$ is the font size (in arcmins) of the letter being tested, $c$ is the probability of a correct answer when guessing randomly, and $k_0$ is the font size at which a patient can start to discern visual information. In an acuity test, we are trying to identify $k_1$: this is the font size at which a patient can see with probability $\tau$, where $\tau$ is some predefined constant ``target probability" specific to the type of eye exam. In this paper we use $\tau=0.80$ which means at font size $k_1$, a patient can correctly guess letters with 80\% probability.
\end{framed}



The Floored Exponential VRF fits the twelve patients with diverse eye diseases and, unlike the logistic function, it has a more believable generative story (a mixture of two processes).
We are currently conducting a large clinical trial that is adding to a growing body of evidence that the VRF is better fit by a Floored Exponential. In the discussion, we explore the possibility that the Floored Exponential mechanism is being misinterpreted as a logistic in other fields as well.

\section{The Stanford Acuity Test (StAT)}

The StAT Test is a novel eye exam based on an improved model of acuity and an intelligent inference process, named after the city it was invented in. StAT employs the Floored Exponential as its VRF and uses likelihood weighted sampling to determine the posterior of $k_1$ given the patient's responses so far. The next letter size to query is then selected by sampling from this posterior. Such an approach balances exploration-exploitation in an optimistic way, in a manner similar to Thompson sampling.  We also include a probability term for the chance that a user ``slips'' and chooses the wrong answer.

\vspace{0.5em} \noindent \textbf{Algorithm.}
We formalize the algorithm as follows. At all times, a StAT digital eye exam keeps track of its belief for the visual acuity of the test taker based on the sequence of answers seen so far $\mathcal{D} = [d_0, d_1, \dots, d_n ]$. Each observation is a tuple $d_i = (x_i, y_i)$ of the size of letter shown to the patient $x_i \in \mathbb{R}^+$ and whether the letter was correctly guessed $y_i \in \{0, 1\}$.
This past data is used both to determine which letter size to query next and also to diagnose the final acuity of the patient at the end of the exam. The StAT algorithm is formally defined in Algorithm \ref{alg:StAT}.


\vspace{0.5em} \noindent \textbf{Computing posterior.}
The continuous distribution for the joint assignment of our two latent variables $k_1$ and $k_0$ given a set of observations $\mathcal{D}$ can be calculated by applying Bayes rule:
\begin{align}
     f(k_1, k_0 | \mathcal{D})
     &\propto f(k_0, k_1) \cdot p(\mathcal{D} | k_0, k_1)\\
     &\propto f(k_1) f(k_0 | k_1) \prod_{i=1}^n p(d_i | k_0, k_1) \label{eqn:posterior_k1_k0},
\end{align}
where $p(d_i | k_0, k_1) = v(x_i, k_0, k_1)$ if $y_i = 1$ (see Floored Exponential box) and
$1 - v(x_i, k_0, k_1)$ otherwise.

\vspace{0.5em} \noindent \textbf{Likelihood Weighting.}
Exact inference of the marginalized posterior of $k_1$ given $\mathcal{D}$ is:
\begin{align*}
    p(k_1 | \mathcal{D}) = \int_{k_0} p(k_0, k_1 | \mathcal{D}) \text{ } d k_0.
\end{align*}
To the best of our knowledge this equation does not have an analytical solution. However using likelihood weighting \cite{shwe1991empirical},  we can sample particles from the joint posterior $f(k_1, k_0 | \mathcal{D})$ given by Equation \ref{eqn:posterior_k1_k0}. We first sample $k_1$ from it's prior and then sample $k_0$ from $p(k_0|k_1)$, weighting the particles by $p(\mathcal{D}|k_1,k_0)$ . We sample a total of 5,000 particles which densely covers the two parameters. After drawing particles from the posterior, the $k_1$ values of those particles represent the distribution $f(k_1 | \mathcal{D})$ and, as such, these particles approximate a soft belief about acuity over the continuous range of possible acuity scores.

We don't discard any particles for a patient between patient queries. After we receive a new datapoint $d_i$, we simply re-weight each particle by multiplying their previous weight by $p(d_i | k_0, k_1)$, using the particle's values for $k_0$ and $k_1$. This makes the computation time of the update step grow linearly with the number of particles and constant with respect to the length of the exam.

\begin{figure}[bt]
    \centering
    \includegraphics[width=0.9\linewidth]{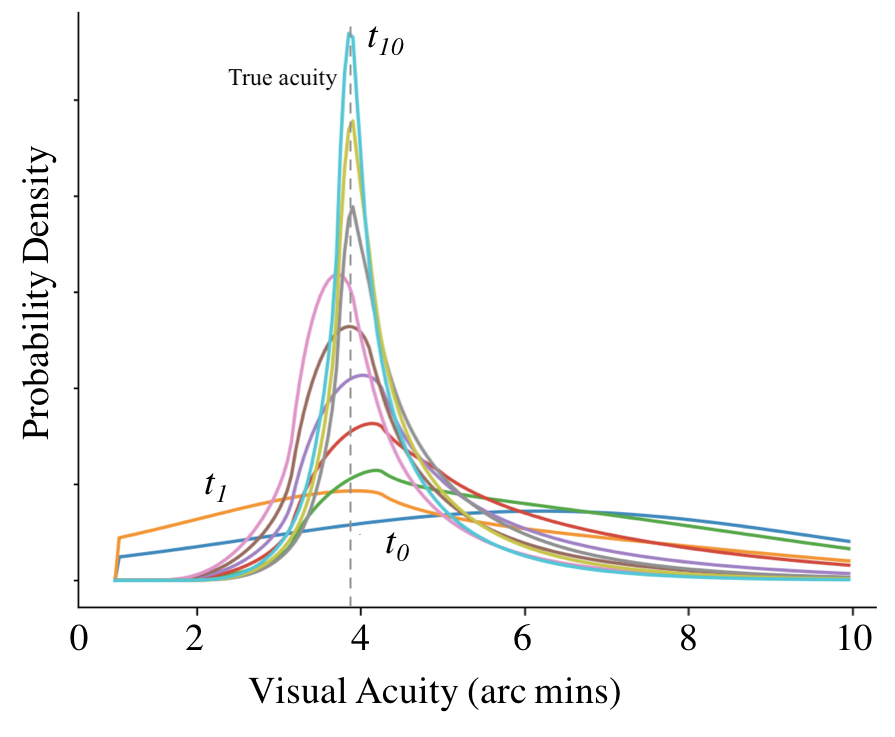}
    \caption{Our model maintains a soft belief about the posterior $p(k_1|d_0, \dots d_i)$ at each timestep $t_i$ in the test.}
    \label{posterior}
\end{figure}

Figure \ref{posterior} shows an example of the posterior distribution for $k_1$ (the statistic for visual acuity) changing over the course of one acuity exam. Initially there is an uncertain belief about the patient's acuity. As the exam progresses, the posterior converges to the true acuity.

\begin{algorithm}[!b]
  \caption{{\footnotesize The Stanford Acuity Test (StAT)}}
  \label{alg:StAT}
  \textbf{Inputs:}
    \begin{itemize}
        \item A patient with an unknown VRF, $p(x; k_0, k_1)$, in the Floored Exponential family.
        \item A length $N$ of maximum questions to ask the patient.
    \end{itemize}

 \textbf{Algorithm: }
    \begin{enumerate}
        \item Inititalise belief of $p(k_1)$ with prior.
        \item For $i = 1, \ldots, N$:
           \begin{enumerate}[label=\roman*)]
                \item Sample $x_i \sim p(k_1 | d_1, \ldots, d_{i-1})$ from current belief of $k_1$.
                \item Query patient at letter size $x_i$ and record whether response was correct as $y_i$. Store $d_i = (x_i, y_i)$.
                \item Update posterior belief of $k_1$ to get  $p(k_1 | d_1, \ldots, d_{i})$.
            \end{enumerate}
        \item Return $\arg \max_{k_1} p(k_1 | d_1, \ldots, d_N)$
    \end{enumerate}
\end{algorithm}

\vspace{0.5em} \noindent \textbf{Prior over $\bm{k_1}$.}  This Bayesian approach requires us to provide a prior probability for $k_1$. Thanks to \citet{bach2006freiburg}, we obtained over a thousand acuity scores of patients. Based on this data, we observed that the log of the acuity score was well fit by a Gumbel distribution. The best-fit prior for the data was $\log k_1 \sim \text{Gumbel}(\mu = -0.1, \beta = 0.3)$. In acknowledgement of the fact that we can't be sure that users of our test will come from the same distribution collected by FrACT, we set our Gumbel prior to be less confident $\log k_1 \sim \text{Gumbel}(\mu = 0.3, \beta = 0.5)$.

 Although we fit a generic prior, if a patient (or doctor) has a belief about the patient's acuity score, they can express that belief via a different Gumbel prior where $\mu$ is the best guess acuity (in LogMAR units) and $\beta$ is a reflection of confidence in the prior. If a user has a reasonable idea of their vision, our acuity algorithm will be quicker and more accurate.

\vspace{0.5em} \noindent \textbf{Slip Probability.} Even if a user can see a letter, they sometimes get the wrong answer because they  ``slip'' and accidentally provide the wrong response or their answer is incorrectly entered. Explicitly modelling this source of uncertainty is as important for a digital eye exam, as it is in traditional tests \cite{cao2008bayesian}.

To account for this slip probability, We replace the VRF $v(x)$ with $v_s(x)$ where $s$ is the slip probability:
\begin{align*}
v_s(x) = s \cdot c + (1-s) \cdot v(x).
\end{align*}
We included this extension after observing that slip mistakes would lead to inaccurate predictions unless explicitly modelled (see noSlip in Table 1).

\vspace{0.5em} \noindent \textbf{Choosing query letter.} An essential step in the intelligence of this algorithm is to decide which next letter size to query the patient. One simple approach would be to query at the most likely MAP estimate of $k_1$ according to the current belief. Although sensible, this method suffers from being overly greedy in its search for the true acuity of the patient, an issue we notice in the performance of this algorithm (see greedyMAP in Table \ref{table:alg_results}).

 The problem with greedily using the current MAP estimate of a distribution comes up often in a different setting in Artificial Intelligence: that of multi-armed bandits problem. Specifically, the Thompson sampling algorithm models the posterior reward distribution of each arm and samples from this distribution rather than picking the most likely value in an effort to balance exploration and exploitation.

We use a similar idea in our algorithm: to determine the next letter size to query the patient, the algorithm samples from its current posterior belief over $k_1$. This means the algorithms is likely to pick exploratory letter sizes at the start, when it is less confident (high variance), and becomes increasingly greedy as its confidence increases. This method is sometimes called Posterior Probability Matching and it is believed to be the way humans make decisions under uncertainty \cite{sabes1996reinforcement,wozny2010probability}.


In contrast to this, the FrACT test uses a purely greedy variance minimization strategy for choosing the next letter size. Specifically, it selects the optotype size that maximizes the likelihood of observation (and thus minimizes the variance in acuity belief). This is a reasonable strategy, but, since the test is $N$ steps long, it suffers from the aforementioned problems and tends to fail at initial exploration of the space.


\section{Experiments}

\begin{figure*}[!ht]
    \centering
    \includegraphics[width=0.95\linewidth]{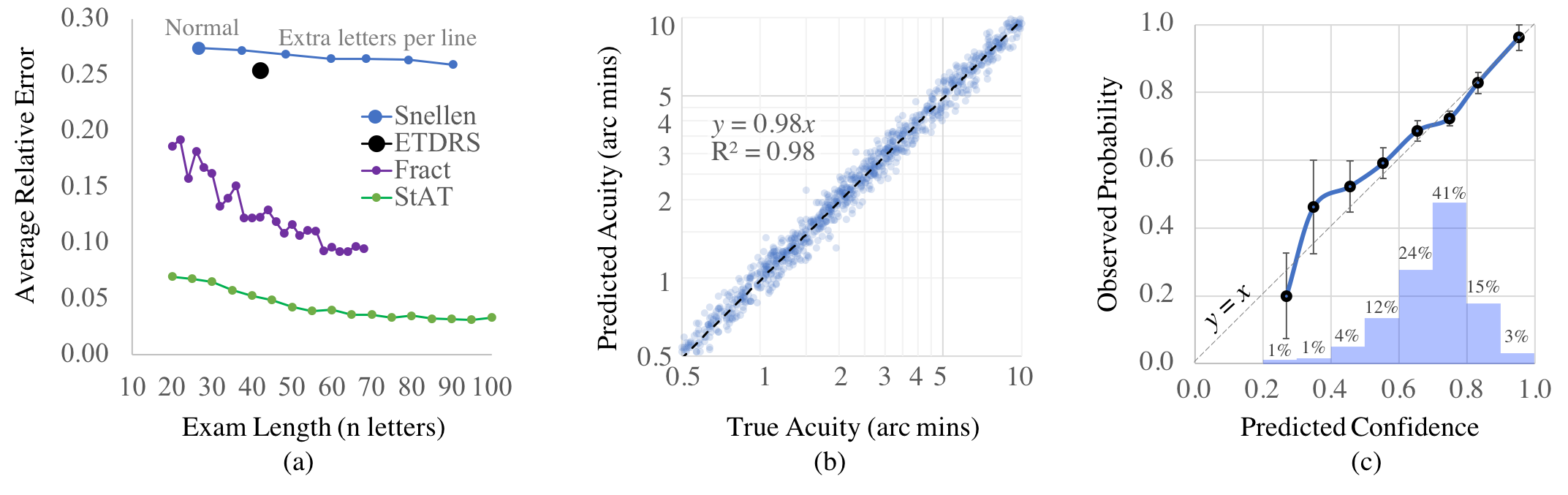}
    \caption{(a) The tradeoff between length of exam and error for the different algorithms. (b) A visualization of the predictions made by StAT. (c) Calibration test: StAT confidences correspond to how often it is correct.}
    \label{fig:row_results}
\end{figure*}

\subsection{Performance Evaluation}

To evaluate the performance of our algorithm with respect to other policies, we simulate patients by sampling parameters for the Floored Exponential VRF, in a manner similar to \citet{shamir2016comparison}. Specifically, for all  experiments we sample 1000 random patients and use them to simulate the performance of each policy. Since we sample the true acuity parameters, we can simulate the exam process and also measure the accuracy of each policy. Acuity scores, $k_1$, are sampled from a high variance Gumbel, with a mode of 2 arcmins. We add a small $s=0.05$ slip probability to responses.

\vspace{0.5em} \noindent \textbf{Measuring error.}
After a virtual acuity test has been run, we have two numbers: the true acuity of the simulated patient, and the acuity that the algorithm diagnosed. From these two numbers we calculate the relative error, which measures the percentage deviation of the prediction from the true acuity.



We use relative error in place of absolute error because of the logrithmic nature of visual acuity. It is generally meaningful to say that a prediction is off by 10\%. In contrast, a prediction which has an absolute error of 1.0 arc mins could be a terrible prediction for a patient with perfect vision (prediction: 20/40, truth: 20/20) but a great prediction for a patient with low vision (prediction: 20/110, truth: 20/100).

\vspace{0.5em} \noindent \textbf{Why not evaluate on humans?}
At first blush, it seems that we should evaluate algorithms on their ability to ``perform well" on real humans. However testing an algorithm on people has a huge downside: there is no gold standard. Imagine an algorithm that runs a visual acuity test on a human. At the end of the test the algorithm declares the patient has an acuity score of 20/42. Was that correct? Unfortunately, there is no way to tell. There is no procedure, invasive or otherwise, that can let you know how well the patient could truly see.

Historically, algorithms were measured on their ability to give the same score on a test and a subsequent retest on the same human. This is a poor measure as it rewards algorithms that make the wrong prediction (as long as that wrong acuity score is repeated). In this case two wrongs shouldn't make a right. To give an example of how bad this can get, an unintelligent algorithm that predicts every individual to have 0 acuity, has a retest rate of 100\%.

Simulations on the other hand can sample a latent true acuity and test if an algorithm can infer that chosen value. As such there \emph{is} a gold standard. For this reason, simulation is becoming the preferred method of evaluation \cite{shamir2016comparison}.
The threat to validity of a simulation is that we have to assume that the simulation is correct. To assuage this concern, algorithms should be tested on a variety of simulation assumptions and substantial effort needs to be put into validating our simulation models. In fact, this need is exactly what led to the discovery of the Floored Exponential.


\subsection{Baseline Acuity Tests}

We use the following baselines and prior algorithms to compare against the StAT algorithm.

\vspace{0.5em} \noindent \textbf{Const Policy.}
This policy always predicts the most common visual acuity in our data i.e. the mode of the visual acuity prior. This serves as a true null model because it doesn't take patient responses into account at all.

\vspace{0.5em} \noindent \textbf{Snellen and ETDRS.}
The Revised 2000 Series ETDRS charts  and the Traditional Snellen Eye Chart  were programmed so that we could simulate their response to different virtual patients. Both exams continue until the user incorrectly answers questions for more than half of the letters on a line. ETDRS has a function for predicted acuity score that takes into account both the last line passed, and how many letters were read on the last line not-passed. Both charts use 19 unique optotypes.

 \vspace{0.5em} \noindent \textbf{FrACT.} We use an implementation of the FrACT algorithm \cite{bach1996freiburg}, with the help of code graciously shared by the original author. We also included the ability to learn the ``$s$'' parameter as suggested by the 2006 paper \cite{bach2006freiburg}, and verified that it improved performance.

\section{Results and Evaluation}

\begin{table}\centering
{
    \begin{tabular}{lcccc}
    \toprule
      & $\mu$ Acuity Error & $\mu$ Test length\\
    \midrule
     Const & 0.536 & 0\\
     Snellen$^\dagger$ & 0.264 & 27\\
     ETDRS$^\dagger$ & 0.254 & 42\\
     FrACT & 0.212 & 20\\
     StAT & \textbf{0.069} & 20\\
     \midrule
      StAT-noSlip & 0.150 & 20\\
      StAT-greedyMAP & 0.132 & 20\\
      StAT-logistic & 0.125 & 20\\
      StAT-noPrior & 0.090 & 20\\
     \midrule
     StAT-goodPrior & \textbf{0.047} & 20\\
     StAT-star & \textbf{0.038} & 63\\
    \bottomrule
    \end{tabular}
}
\caption{Average relative error for each algorithm. Except for Snellen each test was allowed 20 letters. Results are average relative error after 1000 tests. $\dagger$ Snellen and ETDRS used 19 unique optotypes.}
\label{table:alg_results}
\end{table}

The results of the experiments can be seen in Table \ref{table:alg_results}.

\vspace{0.5em} \noindent \textbf{Accuracy and error.} As can be seen from Table \ref{table:alg_results}, the StAT test has substantially less error than all the other baselines. After 20 optotype queries, our algorithm is capable of predicting acuity with an average relative error of 0.069. This prediction is a 74\% reduction in error from our implementation of the ubiquitous Snellen test (average error = 0.276), as well as a 67\% reduction in error from the FrACT test (average error = 0.212). One possible reason for the improvement over FrACT is that the simulations used in our evaluations are based off the Floored Exponential model that StAT uses. However, even when we evaluate StAT on simulations drawn from the FrACT logistic assumption, we still achieve a 41\% reduction in error. The improved accuracy of the StAT algorithm suggests our Bayesian approach to measuring acuity is a fruitful proposal, both because of our introduction of the Floored Exponential as well as our posterior sampling based algorithm for choosing the next letter size to query.

Figure \ref{fig:row_results} (b) visualizes what StAT's small relative error means in terms of predictions. Each point in the plot is a single patient. The x-axis is the true acuity of the patient and the y-axis is the predicted accuracy. We can qualitatively observe that the predictions are often accurate, there are no truly erroneous predictions, and that the exam is similarly accurate for patients of all visual acuities.

Moreover, as seen in Figure \ref{fig:row_results} (a), StAT's significant improvement in error rate holds even when the length of the exam is increased. It is also evident that increasing exam length reduces our error rate: if we increase the exam length to 200 letters, the average error of StAT falls to 0.020. While this is highly accurate, its far too long an exam, even for patients who need to know their acuity to high precision.

\vspace{0.5em} \noindent \textbf{StAT Star Exam.}
Our primary experiments had a fixed exam length of 20 letters. However, since our algorithm models the entire belief distribution over $k_1$, we can run an alternative test that keeps asking the patient queries until it has a 95\% confidence that the relative error is less than  $\epsilon$ $=0.10$. We call this the StAT-star test, and it should be the preferred test for patients who want to have a high confidence in their score.

After running StAT-star 1000 times, 95.1\% of results had error less than 0.10, suggesting that the algorithm's confidence is well calibrated. The exam is longer with an average length of 63 optotypes, but had the lowest average error of all tests: 0.038.

\vspace{0.5em} \noindent \textbf{Improved prior.} We experimentally verified that if a user already had a reasonable understanding of their vision, they could express this as a prior and get more accurate exam results. For example, we saw if a patient was able to guess their vision to within $\pm$ 1 line on a Snellen chart, then the average error of the standard 20 question StAT test would drop to 0.051.

\vspace{0.5em} \noindent \textbf{More optotype choices.} StAT was evaluated using four unique optotype choices (the tumbling-e optotype set). Our algorithm improved slightly as the number of optotype options increased. If we instead use 19-letter optotype options (and thus a guess probability of $c = \sfrac{1}{19}$), error drops to an average error of 0.052.

\vspace{0.5em} \noindent \textbf{Robustness to slip.} Our results proved to be quite invariant to an increase in slip probability, as long as the slip probability was bellow $1/3$. For larger likelihood of slip, our performance started to degrade.


\vspace{0.5em} \noindent \textbf{Importance analysis.} Since our model contributed several extensions to the state of the art, we performed an importance analysis to understand the impact of each individual decision: (1) model slip or noSlip (2) posterior sample or greedyMAP (3) Floored Exponential VRF or logistic (4) gumbel prior or noPrior. For each decision we ran error analysis with that decision ``turned-off". All four decisions had a large increase in error when they were turned-off, suggesting that they were all useful in making a low error test.

When we turned-off the decision to explicitly model accidental ``slips", we had the largest increase in error. While all of our contributions were useful, modelling slips is clearly an important addition.

\vspace{0.5em} \noindent \textbf{Calibrated uncertainty.}
One of the novel abilities of the StAT algorithm is that it can express its confidence in terms of probabilities. To evaluate the reliability of the confidences computed by the StAT test, we plot a calibration curve for the algorithm (see Figure \ref{fig:row_results} (c)). We ran 10,000 experiments of the StAT algorithm: for each run, we recorded both the final predicted value of $k_1$ as well as the probability, according to the algorithm, that $k_1$ was within a relative error of $0.1$ of the true acuity $k_1^*$. We then binned these probabilities and, for all the entries in a bin, computed the empirical fraction of times the algorithm was correct (``empirical success rate"). We compare the predicted confidence to the empirical success rate.

For a perfectly calibrated model, this plot should look like a straight line $y = x$. As we can see in Figure \ref{fig:row_results} (c), the model's confidence is well-calibrated and thus reliable as a measure of uncertainty. The figure also shows that after 20 questions, the algorithm often predicts an 80\% probability that relative error is within 0.1. StAT is not always correct, but unlike previous models it has a calibrated sense for how confident it should be.


\section{Discussion}

The algorithm we have presented in the paper demonstrates a promising approach to measuring the visual acuity that is more accurate while also providing robust notions of uncertainty. In this section, we discuss the implications of this idea, highlighting important limitations and future work.

\subsection{Real World Considerations}
 Although this work has potential for huge impact in diagnosing and treating vision related illnesses, caution must be taken before using this algorithm in a real-word setting.

  \vspace{0.5em} \noindent \textbf{Benefit of variation.}
  In our consideration of eye exams, we did not explicitly design for human fatigue. Being asked the same size question repeatedly is tiring for patients, especially if the letters are difficult to see. Some digital exams mitigate this concern by intentionally giving users ``easy" questions every once-in-a-while. The Monte Carlo sampling aspect of the StAT decision making process naturally leads to a highly dynamic set of letter sizes. We believe this helps keep the test interesting and reduces fatigue.

 \vspace{0.5em} \noindent \textbf{Floored Exponential assumption.}
 One of the biggest assumptions in our paper is that the human VRFs matches the Floored Exponential function. Although we tested this assumption on a number of actual patients with real eye diseases and found promising fits, more clinical trials at a larger scale would be needed to be confident in this facet of the algorithm and to understand if there are certain eye diseases for which it is not the correct parametric form. This same limitation exists in other eye exams as well, for example the ``logistic" assumption built into the FrACT exam, which is used in clinical settings. See Figure 5 for our initial results in this deeper exploration.

 \vspace{0.5em} \noindent \textbf{Peripheral vision.}
 A possible concern for medical practitioners in using a test like StAT involves the role peripheral vision plays in traditional eye exams. According to the literature, checking acuity with single optotypes instead of lines over-estimates true acuity due to an effect known as the crowding phenomenon \cite{LALOR201631}. The scheme discussed in this paper easily extends to showing multiple letters at a time. Presenting single letters has the advantage that it is easy for patients at home to self administer the test (the user interface is more straight forward), however for hospitals we recommend the multiple line version.

 \vspace{0.5em} \noindent \textbf{Convention.}
 Switching to a more accurate system like StAT could require a recalibration of the medical Ophthalmology literature that was built on traditional acuity exams.
 Our results show that current measures of visual acuity are highly susceptible to inaccuracies. Since these measures are used in research when designing appropriate diagnoses and prescriptions, it could be the case that the field has naturally adapted to the errors of traditional chart-based exams.

\subsection{Beyond Eye Exams}
Both of the main contributions in this paper: the VRF and the more intelligent adaptive test, have great potential to contribute to any psychometric test, well beyond vision.

The core idea behind the VRF could extend beyond just visual acuity. In educational Item Response Theory, the probability of a student answering a multiple choice question correctly is currently modelled as a logistic, with the input representing the easiness of the question and the output representing the probability of a correct answer from the student.
The effectiveness of the Floored Exponential function over the logistic function as a model for the acuity function suggests that it may be useful, even for education. Intuitively, the generative story makes sense: when the question is absurdly difficult, the best a student can do is guess. Otherwise, they possess useful information about the question which combines in an exponential manner. Exploring this model in the understanding student responses to questions is an interesting future direction.

Moreover, the sampling inspired algorithm we use is a novel method for the ``computer adaptive test'' problem. In the case of visual acuity testing it was an improvement over the contemporary MLE based exam. Therefore, it may prove to be an improvement over MLE based computer adaptive tests in other domains as well.

\subsection{Future Work}

We hope the ideas here provide a foundations for even further research into improving our ability to diagnose and treat eye related diseases. We outline some seemingly fruitful directions of future research.

\vspace{0.5em} \noindent \textbf{Clinical trials.} An essential next step in demonstrating the usefulness of these ideas is to actually try them on real patients with a series of controlled trials. These experiments would provide insight into the failure modes of our approach as well as other unforeseen factors such as the cognitive load of taking a StAT acuity test. Such research, in conjunction with input from the medical community, could truly transform the ideas in this paper into an actionable reality.


\vspace{0.5em} \noindent \textbf{Smarter letter querying.}
There is potential for investigating a more intelligent way to pick the next letter size based on current belief. One direction we want to explore is proving some optimality bounds on our approach. Another orthogonal investigation would involve learning a policy for picking the next letter size that optimises an objective like minimising test length or maximising confidence.


\section{Conclusion}

Vision-limiting eye diseases are prevalent, affecting billions of people across the world \cite{visionPrevalence}. For patients with serious eye diseases, the ability to finely measure acuity could be a crucial part in early diagnosis and treatment of vision impairment. In this paper, we present a novel algorithm based on Bayesian principles for measuring the acuity of a patient. This algorithm outperforms all prior approaches for this task while also providing reliable, calibrated notions of uncertainty for its final acuity prediction. Our approach is incredibly accurate, easy to implement, and can even be used at home on a computer. With further research and input from the medical community, we hope for this work to be used as a foundation for revolutionising the way we approach visual acuity testing for people around the world.

\subsubsection*{Acknowledgements}
On a personal note, we embarked on this research because several of the authors have experienced eye disease. This work is meant as a ``thank you'' to the many doctors who have helped us keep our vision. We also thank Jason Ford and Ben Domingue for their guidance.

\appendix

\section*{Appendix}

\section{Units of Acuity}\label{apdx:units}

Visual acuity, and optotype size, measure visual angle subtended at the eye by an optotype, in minutes of visual arc. Because the semantic meaning of vision loss is better expressed in a logrithmic space, logMAR, the $\log_{10}$ of the minimum visual angle of resolution, is a popular choice of units. In the Snellen chart, the visual angle is articulated via a fraction in meters (e.g. 6/6) or in feet (e.g. 20/20). The ETDRS chart represents acuity in logMAR units.  In this paper we use visual angle (arcmins) as our unit space and its log, logMAR \cite{westheimer1979scaling}.

Note that the original FrACT paper uses ``decimal" units (1/visual angle) and equation (1) is the FrACT assumption written for visual angle units.

\section{Reparametrising the Floored Exponential}\label{apdx:fexp_reparam}

The Floored Exponential $v(x)$ consists of a floor $c$ and an exponential function, $1 - e^{-\lambda(x - b)}$, parametrised by location $b$ and scale $\lambda$. These parameters capture the full class of Floored Exponential functions but have no intuitive interpretation for eye care providers.

To address this issue, we firstly reparamterise the Floored Exponential with parameter $k_0$, which represents the letter size at which the patient can start to discern information. In other words, it is the value of $x$ where the exponential curve is above the floor of $c$. Normally, the exponential function $1 - e^{-\lambda(x - b)}$ exceeds the floor of $0$ at $x = 0$. To have it exceed the floor of $c$ at $x = k_0$, we scale it by $(1 - c)$, shift it up by $c$, and then shift it right by $k_0$. This gives

\begin{align*}
   v(x) &= \max\left\{c, \  (1 - c)(1 - e^{-\lambda(x - k_0)}) + c\right\} \\
        &= \max\left\{c, \ 1 - (1 - c)e^{-\lambda(x - k_0)}\right\}.
\end{align*}

We next replace $\lambda$ with the parameter $k_1$ which represents the patient's true acuity i.e. the letter size at which $v(k_1) = \tau$, for some fixed constant $\tau \in (c,1]$. We solve for $\lambda$ at this letter size:

\begin{align*}
   &\qquad \quad \tau =  1 - (1 - c)e^{-\lambda(k_1 - k_0)} \\
   &\iff e^{-\lambda(k_1 - k_0)}  =  \frac{1 - \tau}{1-c} \\
   &\iff e^{-\lambda(x - k_0)}  = \left( \frac{1 - \tau}{1-c}
   \right)^{\frac{x-k_0}{k_1 - k_0}}.
\end{align*}

This gives the final equation:
\begin{align*}
   v(x, k_0, k_1) &= \max\left\{c, \ 1 - (1 - c)\left( \frac{1 - \tau}{1-c}
   \right)^{\frac{x-k_0}{k_1 - k_0}}\right\}.
\end{align*}

\

\section{Floored Exponential VRF}\label{apdx:fexp_vrfs}
\begin{figure}[!ht]
    \centering
    \includegraphics[width=0.40\linewidth]{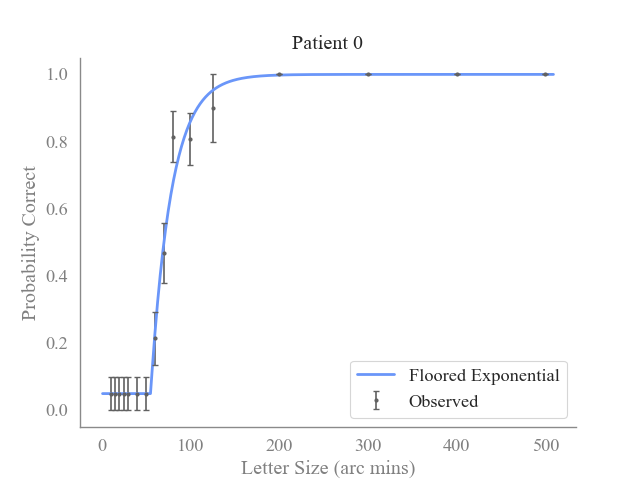}
    \includegraphics[width=0.40\linewidth]{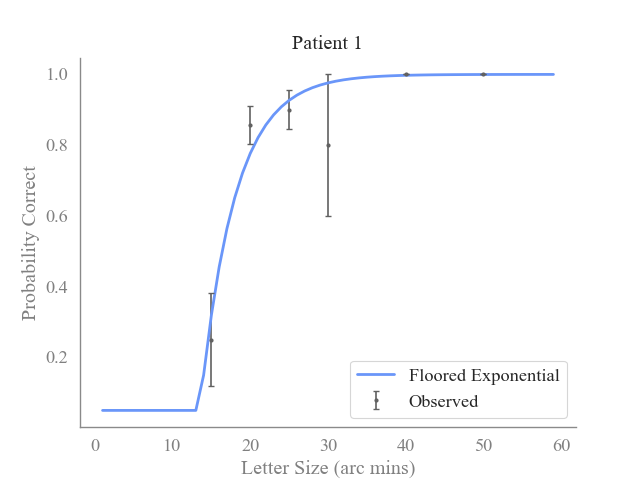}
    \includegraphics[width=0.40\linewidth]{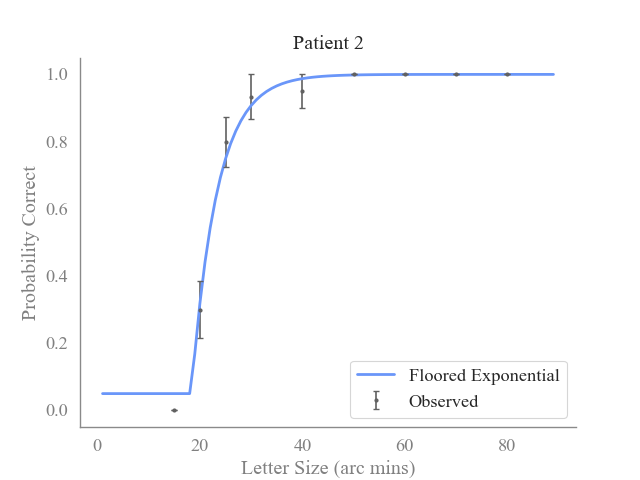}
    \includegraphics[width=0.40\linewidth]{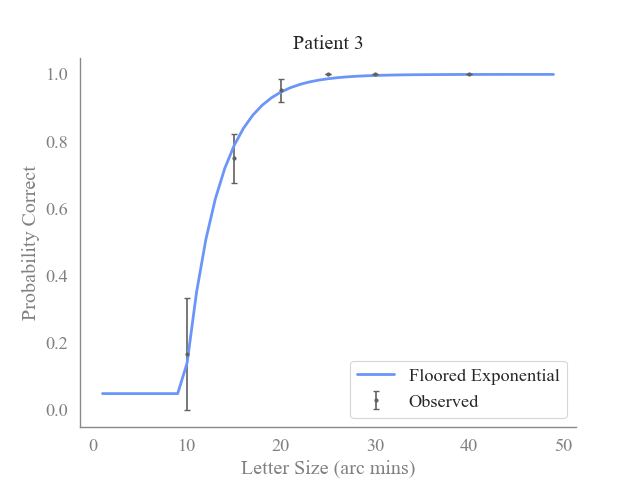}
    \includegraphics[width=0.40\linewidth]{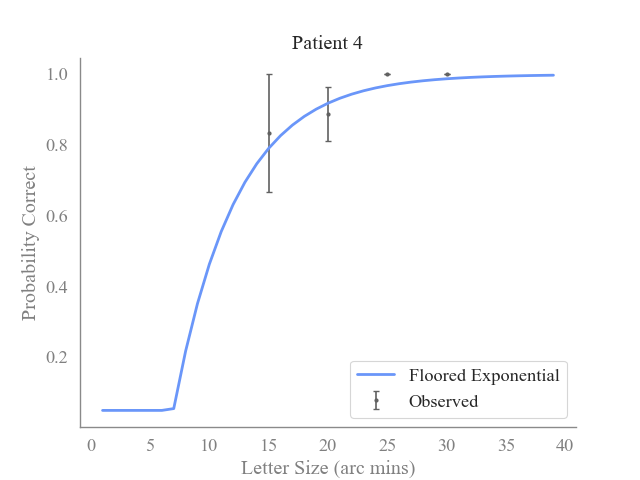}
    \includegraphics[width=0.40\linewidth]{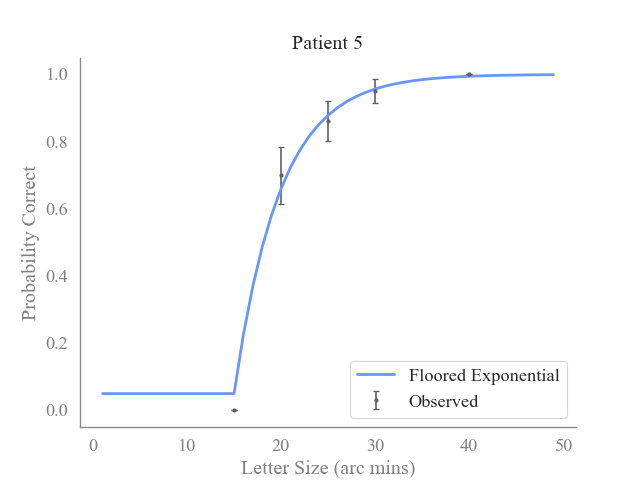}
    \includegraphics[width=0.40\linewidth]{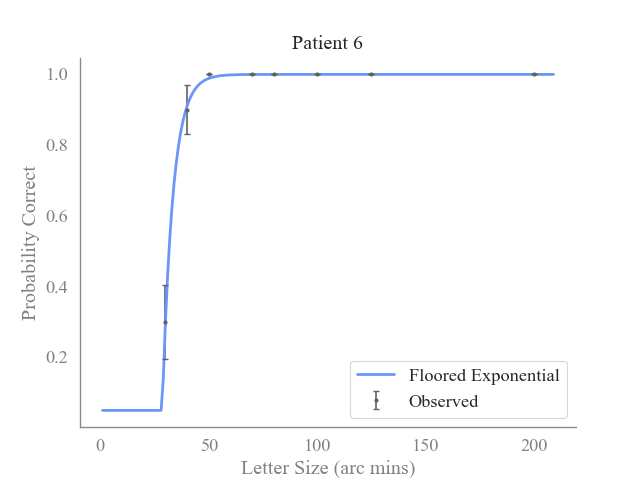}
    \includegraphics[width=0.40\linewidth]{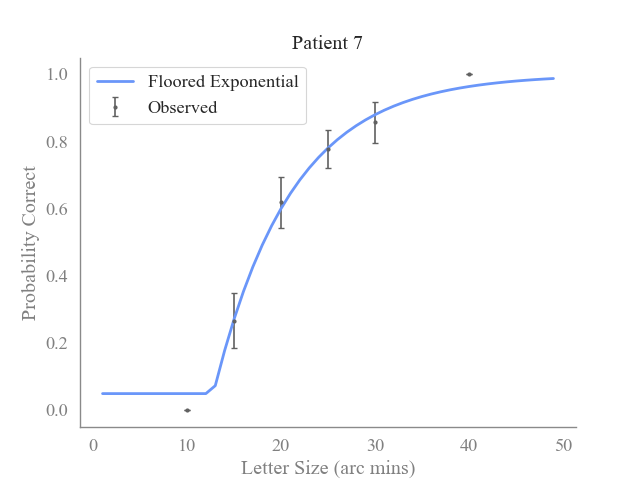}
    \caption{As part of our ongoing research we are verifying that the Floored Exponential fits different patients with different eye diseases. Here are the curves from eight patients from our study.}
    \label{fig:patient_fits}
\end{figure}




\bibliography{main}
\bibliographystyle{aaai}
\end{document}